%% file: main.tex
\documentclass{article}

\usepackage[dblblindworkshop,final]{neurips2026}
\workshoptitle{World Models for High-Stakes Health}

\usepackage[utf8]{inputenc}
\usepackage[T1]{fontenc}
\usepackage{microtype}
\usepackage{amsmath}
\usepackage{amssymb}
\usepackage{amsfonts}
\usepackage{nicefrac}
\usepackage{graphicx}
\usepackage{xcolor}
\usepackage{booktabs}
\usepackage{multirow}
\usepackage{tabularx}
\usepackage{threeparttable}
\usepackage{adjustbox}
\usepackage{siunitx}
\usepackage{enumitem}
\usepackage{hyperref}
\usepackage{url}
\usepackage[percent]{overpic}
\setcitestyle{numbers}
\setcitestyle{numbers,square,comma}

\title{
Block Diffusion for Longitudinal EHR:
Accuracy, Efficiency, and Representation Trade-offs
}

\author{%
  Ashvin Gupta$^{1}$ \quad Joshua Placidi$^{1}$ \quad Brendan Delaney$^{1}$ \quad Alessandra Russo$^{1}$ \\[0.5em]
  $^{1}$Imperial College London \\
}

\begin{document}

\maketitle

\vspace{-10pt}
\input{text/abstract}

\input{text/introduction}

\input{text/methods}
\input{text/results}
\input{text/discussion}


\bibliographystyle{unsrtnat}
\bibliography{references}

\clearpage
\appendix
\input{text/appendix}

\end{document}

%% file: text/abstract.tex
\begin{abstract}
Autoregressive (AR) models have proved effective for zero-shot prediction from longitudinal electronic health records, but impose strictly sequential generation on trajectories containing temporally clustered and interdependent clinical events. We investigate block-based masked diffusion (BD) as an alternative on MIMIC-IV under matched tokenisation, context length, model capacity, and evaluation conditions. Across block sizes, BD exhibits a clear quality--efficiency trade-off, with small-block configurations retaining much of the predictive performance of AR while enabling substantially faster inference. A representative $B{=}8$ configuration achieves approximately $1.43\times$ faster evaluation. Although generative rollout remains below AR in most settings, frozen linear probes on BD and AR representations achieve near-identical performance across clinical prediction tasks, indicating that BD learns comparably informative clinical representations. Increasing the denoising budget yields diminishing returns, while alternative decoding strategies provide limited gains. Overall, our results show that block diffusion is a viable and computationally attractive approach to longitudinal EHR modelling, with further improvements in iterative decoding offering a promising route toward closing the remaining generative performance gap.
\end{abstract}

%% file: text/introduction.tex
\section{Introduction}
Foundation models for structured electronic health record (EHR) data have predominantly adopted autoregressive (AR) sequence modelling. By generating possible future patient trajectories, these models have demonstrated strong zero-shot performance across clinical prediction tasks, in some settings approaching or exceeding task-specific supervised baselines \cite{Rajamohan2025FoundationScale,Renc2024ZeroTransformer,Waxler2025GenerativeScale,Shmatko2025LearningTransformers}. However, AR modelling imposes a strict left-to-right factorisation with each event generated conditionally on all preceding events before subsequent events can be considered. Longitudinal EHRs are only partly sequential in this sense. 
Clinical activity often occurs in temporally concentrated bursts, making groups of nearby events a natural target for partially parallel generation (Appendix \ref{decoding_intervetions}), and in our MIMIC-IV data, 47.4\% of recorded timestamps contain at least eight tokens (Appendix \ref{MIMIC-IV tokens}).
This strict factorisation also carries an inference cost, as token-by-token decoding limits parallelism and requires sequential computation for each generated event. This motivates generative objectives that relax strict next-token prediction and instead allow groups of clinical events to be modelled jointly.

Discrete diffusion language models provide one such alternative by learning to reconstruct corrupted tokens through iterative denoising \cite{Sahoo2024SimpleModels}. 
These models have shown strong generative and representation-learning performance in other structured scientific domains, including protein sequence modelling and molecular graph generation \cite{Wang2024DiffusionLearners,Vignac2023DiGress:Generation}. 
Block diffusion (BD) introduces an AR factorisation across blocks while denoising tokens jointly within each block, enabling partial parallelism while retaining causal conditioning on previously generated context \cite{Arriola2025BlockModels}.
Prior work has shown a quality--parallelism trade-off: larger blocks allow more parallel generation but generally degrade likelihood and generation quality, whereas smaller blocks more closely approach AR performance. Discrete diffusion has previously been applied to EHRs for synthetic data generation \cite{Tian2024ReliableModels, Han2024GuidedGeneration}, but its behaviour relative to strong AR foundation models for zero-shot clinical prediction remains less well understood.

In this work, we conduct a controlled empirical comparison between AR and block-diffusion modelling of tokenised MIMIC-IV patient trajectories. We ask three questions: (1) how do block size and denoising budget affect the trade-off between predictive performance and inference efficiency; (2) when diffusion rollout underperforms AR, does the gap arise from poorer learned representations or from the iterative generation process; and (3) can alternative decoding strategies recover performance lost during blockwise generation?

Our contributions are:
\begin{enumerate}
    \item We show that block diffusion is a competitive alternative for longitudinal EHR prediction, achieving rollout performance close to AR while providing up to a $1.43\times$ evaluation speedup in our primary operating regime.
    \item We show that AR and block-diffusion models encode similarly predictive clinical information, with frozen linear probes achieving near-identical downstream performance across tasks despite a larger gap during generative rollout.
    \item We provide a systematic study of how block size and denoising budget affect the quality--efficiency trade-off of block diffusion on EHR sequences, and find that alternative decoding strategies are not sufficient to close the remaining performance gap to AR.
\end{enumerate}

%% file: text/methods.tex
\section{Methods}

\subsection{Models and generative prediction}
We represent each longitudinal EHR as a discrete token sequence $x=(x_1,\ldots,x_L)$, where each token corresponds to a clinical event, quantised measurement, demographic attribute, or discretised time interval from a vocabulary $\mathcal{V}$.
Both AR and BD models operate on this same temporal sequence.

We compare an AR language model with a capacity-matched BD model. The AR baseline is a GPT-2--style decoder with causal self-attention \cite{Radford2019LanguageLearners}, trained using standard next-token cross-entropy:
\begin{equation}
\mathcal{L}_{\mathrm{AR}}
=
-\sum_{i=1}^{L-1}
\log p_{\theta}(x_{i+1}\mid x_{\leq i})
\end{equation}

For BD, we follow BD3-LM \cite{Arriola2025BlockModels}. The trajectory is partitioned into blocks, with causal conditioning across blocks and masked diffusion within each active block. For a clean block $x_0$, we sample $t\sim\mathcal{U}(0.001,0.999)$ and obtain $x_t$ by retaining each token independently with probability $\alpha_t=\cos(\pi t/2)$ and replacing it with \texttt{[MASK]} otherwise. The model is trained using the continuous-time masked diffusion objective:
\begin{equation}
\mathcal{L}_{\mathrm{MDLM}}
=
-\mathbb{E}_{x_0,t,x_t}
\left[
\frac{-\alpha_t'}{1-\alpha_t}
\sum_{i=1}^{L}
\mathbf{1}\!\left[x_t^i=\mathrm{[MASK]}\right]
\log p_{\theta}
\left(x_0^i \mid x_t,x^{(<k)}\right)
\right]
\end{equation} 
where $x^{(<k)}$ denotes the clean preceding blocks. Independent noise levels are sampled across blocks.




At inference time, both models use KV caching \cite{Pope2022EfficientlyInference}. AR generates token by token, whereas block diffusion initialises each new block as fully masked, iteratively denoises it, and then commits the completed block to the cache. 
For zero-shot clinical prediction, generation continues until the model predicts one of the preconfigured end tokens, or a maximum number of generation steps is reached.
We perform 20 Monte Carlo rollouts per patient and use the fraction in which the positive endpoint occurs first as the prediction score. We report AUROC, AUPRC, and wall-clock evaluation time.

\subsection{Data and experiments}

We use MIMIC-IV v2.2 \cite{Johnson2024MIMIC-IV} and follow the tokenisation framework proposed in \citet{Renc2024ZeroTransformer}, with further details in the paper.
We evaluate ICU mortality, ICU readmission, ICU admission, and hospital mortality.
To isolate the effect of the generative formulation, AR and diffusion models use identical tokenisation, data splits, and context length of 1024, with matched capacity: 6 transformer layers, hidden dimension 768, 12 attention heads, rotary positional embeddings, and the same optimizer schedule. We train separate BD models with block sizes $B\in\{2,4,8,16,32,64,128\}$ and retain ICU mortality as the primary task for ablations. 

We first study how within-block parallelism affects predictive performance and inference efficiency. For the primary block-size comparison, we set the number of reverse steps to $S=B/2$, fixing the nominal denoising budget at $S/B=0.5$ denoiser evaluations per generated token while varying the degree of joint within-block prediction. 
To separate block size from the amount of iterative refinement, we perform fixed-block denoising sweeps for $B=4$, $B=8$, and $B=128$. These experiments test whether additional reverse steps can recover performance lost at larger block sizes.

We next test whether differences in generative performance are already reflected in the learned representations. We freeze the AR and $B=8$ diffusion backbones and fit linear classifiers to their final-token hidden representations for all four clinical outcomes. This removes the generative decoding procedure and directly measures the predictive information encoded by each model. Mean pooling yielded lower performance for both models.

Our default BD decoder uses ancestral random unmasking according
to the reverse masking schedule. Let $t>s$ denote consecutive reverse times,
with $\alpha_t$ the token-retention probability at time $t$. Each currently
masked position is revealed with probability
$p_{\mathrm{unmask}}
=\frac{\alpha_s-\alpha_t}{1-\alpha_t}.$
At $B=8$, $S=4$, we additionally evaluate two inference-time alternatives.
\emph{Confidence-based unmasking} preserves the same expected reveal budget:
if $n_{\mathrm{mask}}$ positions remain masked, we reveal
$k \approx n_{\mathrm{mask}}p_{\mathrm{unmask}}$ positions, selecting those
whose sampled token receives the highest model probability
\cite{Ye2025DreamModels,Chang2022MaskGIT:Transformer}.
\emph{ReMDM-style remasking} allows previously revealed positions to be masked
and reconsidered during subsequent reverse steps \cite{Wang2026RemaskingScaling}.
On non-final steps, revealed positions are remasked independently with probability
$\sigma
=
\min\left(
\eta,
\frac{1-\alpha_s}{\alpha_t}
\right)$,
where $\eta$ controls the maximum remasking rate and we evaluate
$\eta\in\{0.05,0.1,0.2\}$. 

%% file: text/results.tex
\section{Results}
\begin{figure}[t]
    \centering
    \begin{minipage}[t]{0.49\linewidth}
        \centering
        \includegraphics[width=\linewidth]{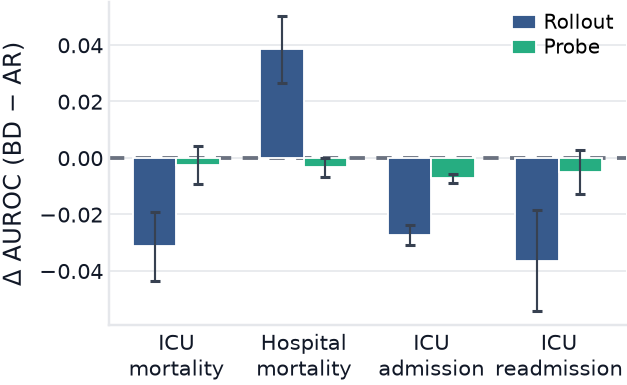}
    \end{minipage}
    \hfill
    \begin{minipage}[t]{0.49\linewidth}
        \centering
        \includegraphics[width=\linewidth]{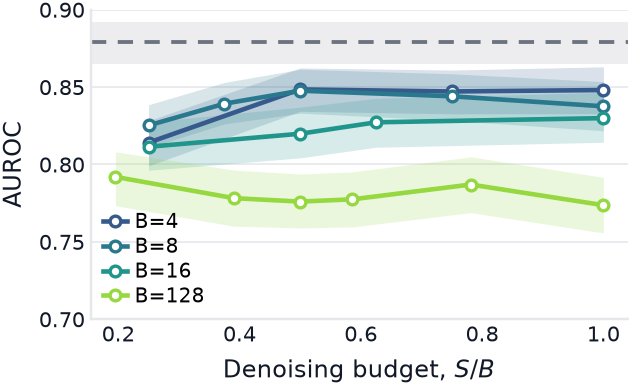}
    \end{minipage}
    
    \vspace{0.4em}

    \begin{minipage}[t]{0.49\linewidth}
        \centering
        \includegraphics[width=\linewidth]{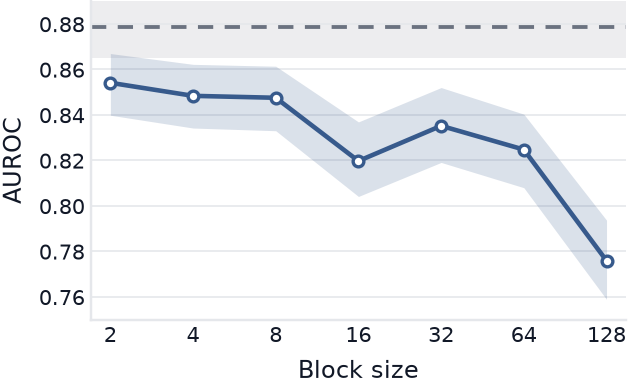}
    \end{minipage}
    \hfill
    \begin{minipage}[t]{0.49\linewidth}
        \centering
        \includegraphics[width=\linewidth]{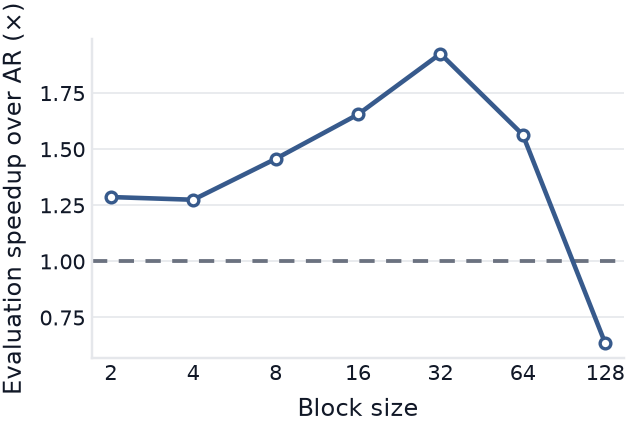}
    \end{minipage}

    \vspace{-0.4em}
    \caption{
    Quality, efficiency, and representation trade-offs of block diffusion. \textbf{Top-left:} BD–AR AUROC difference under rollout and linear probing. \textbf{Top-right:} AUROC across denoising budgets. \textbf{Bottom:} ICU mortality AUROC and evaluation time across block sizes at \(S/B=0.5\). Dashed lines denote AR.
    }
    \label{fig:main_results}
    \vspace{-2em}
\end{figure}

BD provides competitive generative performance across the four downstream tasks, although AR remains stronger on three of four outcomes (Figure~\ref{fig:main_results}). Hospital mortality is the exception, but an increase in AUROC comes at the cost of AUPRC (Appendix \ref{app:hospital_mortality_scores}). Importantly, this gap largely disappears when evaluating the learned representations directly with frozen linear probes achieving near-identical AUROC for AR and BD across tasks. This indicates that BD captures similarly predictive clinical information, and suggests that much of the remaining gap emerging during rollout rather than representation learning.

Under a fixed denoising budget, smaller blocks provide the strongest quality--efficiency trade-off. 
The BD configuration of $B=8$, $S=4$ remained 
approximately $0.03$ AUROC below AR, while achieving a $1.43\times$ evaluation speedup. Performance generally declines at larger block sizes, and very large blocks eventually lose the latency advantage despite requiring fewer sequential forward passes, reflecting the increased cost of each blockwise denoising step.

Increasing the number of denoising steps initially improves predictive performance, but gains saturate beyond a moderate budget while evaluation cost continues to rise. Alternative decoding strategies do not materially change this picture. ReMDM-style revision produces only small, metric-dependent changes, while confidence-based unmasking degrades AUROC (Appendix~\ref{decoding_intervetions}).

%% file: text/discussion.tex
\section{Discussion}

We hypothesised that masked diffusion might be well suited to longitudinal EHR modelling because clinical activity often occurs in temporally concentrated bursts, where the exact ordering of nearby events may be less informative than their joint occurrence. Our results support block diffusion as a practical alternative to strictly autoregressive generation. Small-block BD configurations retain much of the predictive performance of AR while providing faster evaluation, with up to a $1.43\times$ speedup in our primary operating regime. This benefit is not monotonic with block size, however. Large blocks reduce sequential depth but increase per-step compute, producing a U-shaped latency curve and reproducing the quality--parallelism trade-off observed in prior BD work \cite{Arriola2025BlockModels}.

Increasing the denoising budget improves performance initially, but the gains plateau, suggesting that insufficient iterative refinement alone does not explain the remaining gap to AR. More strikingly, this discrepancy is largely absent under frozen linear probing, where AR and BD representations achieve near-identical predictive performance across downstream tasks. This suggests that BD learns similarly predictive clinical representations, and that much of the remaining difference arises during iterative generation rather than representation learning.

Our decoding ablations further support this interpretation. Confidence-based unmasking \cite{Ye2025DreamModels,Chang2022MaskGIT:Transformer} degrades performance, consistent with evidence that it can prematurely commit locally easy predictions before resolving dependencies important to the global trajectory \cite{Kim2026TheModels}. ReMDM-style revision \cite{Wang2026RemaskingScaling} enables iterative refinement but provides no meaningful gain. 
These results suggest that the performance gap is unlikely to be resolved through revision of token commitment alone.

One possible source of this mismatch is the use of a fixed block size. Each block must resolve a predetermined number of positions even though clinically meaningful event groups need not align with these boundaries. This may be particularly relevant for EHR trajectories, where the number of events associated with a clinical episode varies substantially. Variable-length diffusion generation therefore represents a natural direction for future work \cite{Havasi2025EditOperations,Wu2026DreamOn:Canvas,Kim2025Any-OrderDiffusion}.

More generally, EHR structure may motivate diffusion formulations that differ from those developed for natural language. One possibility is to generate events between successive time markers jointly, treating temporally co-occurring events as unordered or partially ordered sets. Alternative corruption processes, including uniform discrete diffusion \cite{Schiff2025SimpleModels,Sahoo2025TheDuality}, may also be worth exploring because they require the model to distinguish corrupted clinical events from plausible clean ones rather than relying exclusively on a dedicated mask state.

Finally, generative endpoint prediction captures only one use of diffusion models for EHRs. Their ability to condition on partially observed sequences could support infilling missing events, reconstructing incomplete records, or exploring counterfactual trajectories. Overall, our results suggest that block diffusion is a promising and computationally attractive framework for longitudinal EHR modelling, while also showing that improved decoding remains important for closing the residual gap to AR zero-shot prediction.

%% file: text/appendix.tex
\section{Appendix}

\subsection{Responsible Use}
This study evaluates block diffusion for retrospective prediction from longitudinal electronic health records and does not establish suitability for clinical deployment. Experiments are limited to MIMIC-IV, and performance may not generalise across institutions, patient populations, or care settings. Generative predictions may exhibit uncertainty, miscalibration, and clinically consequential errors. Before any clinical application, models would require external validation, subgroup fairness assessment, calibration analysis, and prospective evaluation under appropriate clinical oversight. Outputs should not be used to guide patient care without such validation and human supervision and as such, this work should be used for research purposes only.

\subsection{Acknowledgements}
This work is supported by UK Research and Innovation (UKRI AI Centre for Doctoral Training in Digital Healthcare grant number EP/Y030974/1), NIHR Imperial Biomedical Research Centre and the NIHR Invention for Innovation grant ``Artificial Intelligence to support cancer early diagnosis in general practice" (NIHR207533). 

\subsection{Decoding with AR vs BD}
\label{Decoding_appendix}
\begin{figure}[h]
    \centering
    \includegraphics[width=\textwidth]{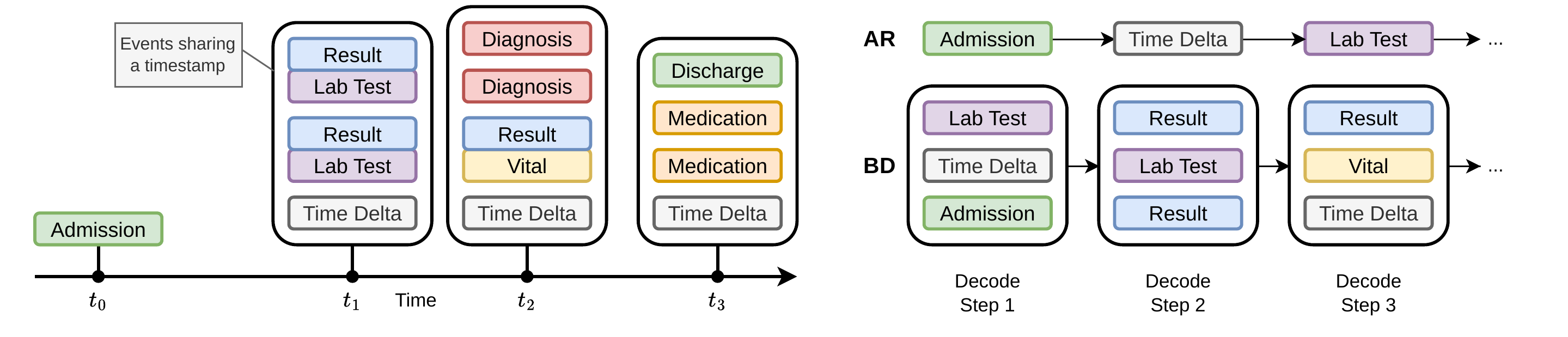}
    \caption{\textbf{Illustration of AR and block-diffusion decoding for longitudinal EHR trajectories.} Clinical events frequently occur in temporally clustered groups. AR generates each event sequentially, whereas BD preserves causal generation across blocks while jointly denoising multiple positions within the active block.}
    \label{fig:ehr_timeline}
\end{figure}

\subsection{MIMIC-IV tokens}
\label{MIMIC-IV tokens}
\begin{figure}[h]
    \centering
    \includegraphics[width=0.5\linewidth]{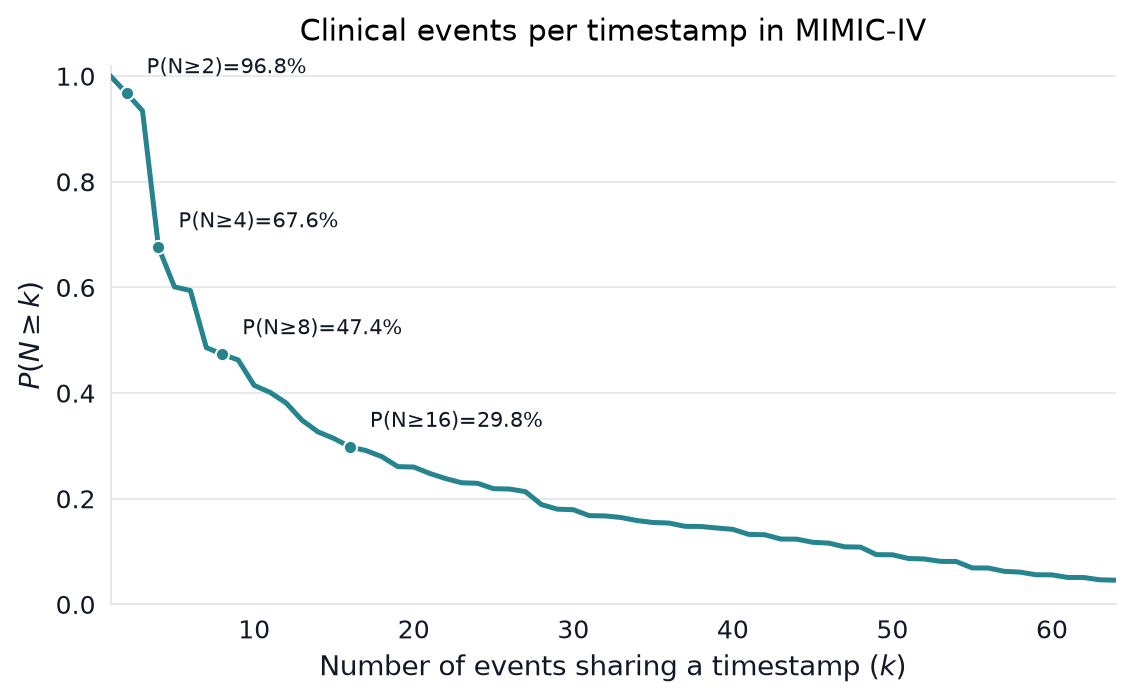}
    \caption{
    \textbf{Clinical event multiplicity in MIMIC-IV.} Survival distribution of the number of non-time-gap event tokens sharing a recorded timestamp in the ETHOS representation.}
    \label{fig:timepoints_survival}
\end{figure}

\begin{table}[h]
\centering
\caption{Multiplicity of clinical event tokens sharing a timestamp in the MIMIC-IV / ETHOS training set. $N$ is the number of non-time-gap event tokens at a timestamp.}
\label{tab:events_per_timestamp}
\begin{tabular}{lc}
\toprule
Statistic & Value \\
\midrule
Number of timestamps & 23\,274\,367 \\
Mean events per timestamp & 16.16 \\
Median & 6 \\
25th / 75th percentile & 3 / 20 \\
$P(N > 1)$ & 96.8\% \\
$P(N \geq 4)$ & 67.6\% \\
$P(N \geq 8)$ & 47.4\% \\
$P(N \geq 16)$ & 29.8\% \\
\bottomrule
\end{tabular}
\end{table}

\subsection{Block diffusion objective and sampling}
\label{app:block_diffusion}

Let a tokenised EHR trajectory be $x=(x_1,\ldots,x_L)$, partitioned into $K$ contiguous blocks $x=(x^{(1)},\ldots,x^{(K)})$ of size $B$. Following BD3-LM \cite{Arriola2025BlockModels}, the sequence likelihood factorises autoregressively across blocks,
\begin{equation}
p_\theta(x)
\prod_{k=1}^{K}
p_\theta!\left(x^{(k)}\mid x^{(<k)}\right),
\end{equation}
where $x^{(<k)}$ denotes the completed preceding blocks. Within each active block, tokens are modelled jointly using masked discrete diffusion.

For a clean block $x_0^{(k)}$, we sample a continuous diffusion time $t\sim\mathcal{U}(\epsilon,1-\epsilon)$ with $\epsilon=0.001$, using the cosine retention schedule
\begin{equation}
\alpha_t=\cos\left(\frac{\pi t}{2}\right).
\end{equation}
Each token is independently retained with probability $\alpha_t$ and otherwise replaced by the absorbing mask state $m=\mathrm{[MASK]}$,
\begin{equation}
q(x_t^i\mid x_0^i)
\alpha_t,\delta_{x_0^i}
+
(1-\alpha_t),\delta_m.
\end{equation}
Noise levels are sampled independently across blocks. The denoiser predicts the original tokens in the corrupted active block while conditioning on the clean preceding blocks. For absorbing-state masked diffusion, the continuous-time objective reduces to the weighted masked-language-modelling loss \cite{Sahoo2024SimpleModels},
\begin{equation}
\mathcal{L}_{\mathrm{MDLM}}^{(k)}
-\mathbb{E}*{t,x_t^{(k)}}
\left[
\frac{-\alpha_t'}{1-\alpha_t}
\sum*{i:x_t^{(k),i}=m}
\log
p_\theta
\left(
x_0^{(k),i}
\mid
x_t^{(k)},x^{(<k)}
\right)
\right].
\end{equation}
The total objective averages this loss across blocks. During training, the active block is observed only in corrupted form, while completed blocks provide clean causal context, matching the conditioning structure available during KV-cached blockwise generation.

At inference time, the next block is initialised entirely as \texttt{[MASK]} and denoised over reverse times $1=t_S>\cdots>t_0=0$. For a reverse transition from $t$ to $s<t$, a currently masked position is revealed with probability
\begin{equation}
p_{\mathrm{unmask}}(s,t)
\frac{\alpha_s-\alpha_t}{1-\alpha_t}.
\end{equation}
Selected positions sample their token identity from
$p_\theta(x_0^i\mid x_t^{(k)},x^{(<k)})$, while unselected positions remain masked. On the final step all remaining masks are resolved, after which the completed block is committed to the KV cache and generation proceeds to the next block.

For confidence-based decoding \cite{Chang2022MaskGIT:Transformer,Ye2025DreamModels}, we preserve approximately the same reveal budget,
\begin{equation}
k_{\mathrm{reveal}}
\approx
n_{\mathrm{mask}},p_{\mathrm{unmask}}(s,t),
\end{equation}
but commit the $k_{\mathrm{reveal}}$ positions whose sampled token receives the highest model probability. ReMDM-style revision \cite{Wang2026RemaskingScaling} instead allows already revealed positions to be reconsidered. On non-final reverse steps, revealed positions are remasked independently with probability
\begin{equation}
\sigma(s,t)
\min\left(
\eta,
\frac{1-\alpha_s}{\alpha_t}
\right),
\end{equation}
where $\eta\in{0.05,0.1,0.2}$ controls the maximum remasking rate.

For a block size $B$ and $S$ denoising steps, we use $S/B$ as the nominal number of sequential denoiser evaluations per generated token. Our primary block-size comparison sets $S=B/2$, giving $S/B=0.5$. This is a normalisation of sequential decoding depth rather than exact compute: the wall-clock cost of each forward pass still depends on the active block size.

\subsection{Hospital-mortality AUROC versus AUPRC}
\label{app:hospital_mortality_scores}

Hospital mortality exhibits an unusual divergence between discrimination
metrics. BD (\(B{=}8\), \(S{=}4\)) outperforms AR in rollout AUROC
(\(\Delta{=}+0.039\), 95\% CI \([+0.027,+0.050]\)), but substantially
underperforms in AUPRC
(\(\Delta{=}{-}0.102\), 95\% CI \([{-}0.124,{-}0.080]\)).
In contrast, last-token linear probes on the same checkpoints are nearly
identical (AR \(0.936\), BD \(0.933\)), suggesting that this discrepancy
arises during generative scoring rather than from a difference in
representation quality.

For each patient, the rollout score
\(\hat p_{\mathrm{death}}\) is the fraction of 20 Monte Carlo trajectories
in which \texttt{MEDS\_DEATH} is generated before the competing endpoint.
The two models produce markedly different score distributions.
AR is conservative, assigning a score of zero to \(23.7\%\) of true deaths
and \(86.7\%\) of survivors. These missed deaths are therefore tied with a
large mass of score-zero survivors, limiting pairwise ranking performance.
However, the relatively few patients receiving high AR scores are precise,
with precision@$100$ of \(0.74\).

BD produces a less sparse score distribution. Only \(2.5\%\) of true deaths
receive a score of zero, and the mean death score increases from \(0.242\)
under AR to \(0.353\). This improves separation between deaths and survivors
and consequently increases AUROC. However, BD also assigns nonzero death probability
to \(60.9\%\) of survivors, compared with only \(13.3\%\) for AR. This
increase in false-positive mass reduces precision, particularly among the
highest-ranked predictions (precision@$100$ \(0.47\)), explaining why
higher AUROC is accompanied by substantially lower AUPRC.

Table~\ref{tab:hospital_mortality_scores} summarises the resulting
instance-level score distributions on the shared test split
(\(n{=}55{,}353\), \(1{,}135\) deaths).

\begin{table}[h]
\centering
\caption{Hospital-mortality rollout scores on the test set.
AUROC/AUPRC brackets are \(95\%\) instance-bootstrap CIs.
Score statistics are Monte Carlo \(P(\texttt{MEDS\_DEATH})\) over \(20\) samples.}
\label{tab:hospital_mortality_scores}
\begin{tabular}{lcc}
\toprule
 & AR & BD (\(B{=}8\), \(S{=}4\)) \\
\midrule
AUROC & \(0.844\) \([0.831, 0.858]\) & \(0.883\) \([0.873, 0.892]\) \\
AUPRC & \(0.306\) \([0.276, 0.335]\) & \(0.204\) \([0.183, 0.227]\) \\
\text{Precision@100} & \(0.74\) & \(0.47\) \\
Mean score (deaths / survivors) & \(0.242\) / \(0.015\) & \(0.353\) / \(0.090\) \\
Score \(=0\) (deaths / survivors) & \(23.7\%\) / \(86.7\%\) & \(2.5\%\) / \(39.1\%\) \\
\bottomrule
\end{tabular}
\end{table}

\subsection{Quality--efficiency operating points}
\begin{figure}[h]
    \centering
    \includegraphics[width=0.5\linewidth]{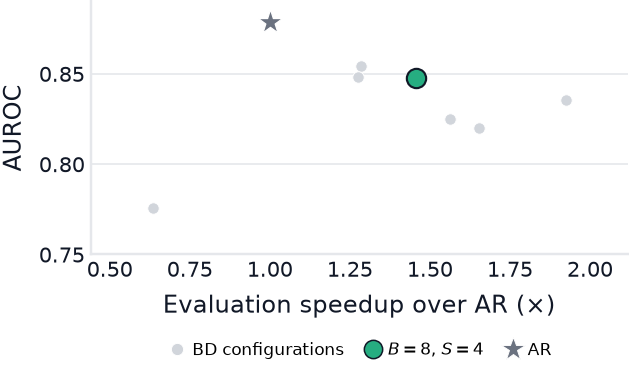}
    \caption{\textbf{Predictive performance versus evaluation speed on ICU mortality.}
Each point denotes a block-diffusion configuration with a particular block size and denoising budget. The x-axis shows evaluation speedup relative to the autoregressive (AR) baseline, while the y-axis shows AUROC. The highlighted $B=8$, $S=4$ configuration provides a representative operating point, achieving approximately $1.43\times$ faster evaluation while retaining much of the predictive performance of AR.}
    \label{fig:placeholder}
\end{figure}




\clearpage
\subsection{AUPRC results}

\begin{figure}[h]
    \centering

    \begin{minipage}[h]{0.32\linewidth}
        \centering
        \includegraphics[width=\linewidth]
        {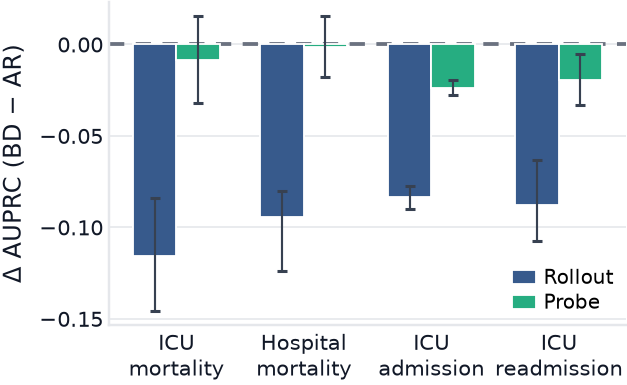}
    \end{minipage}
    \hfill
    \begin{minipage}[h]{0.32\linewidth}
        \centering
        \includegraphics[width=\linewidth]
        {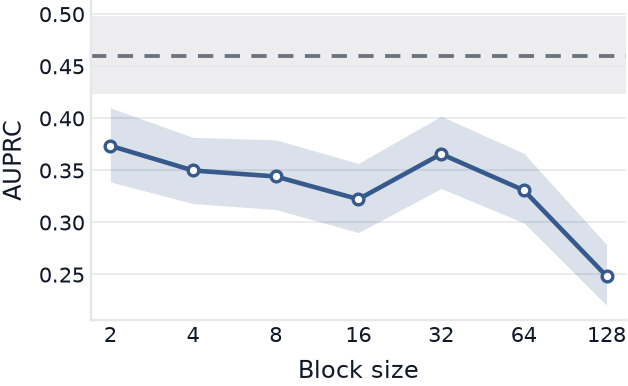}
    \end{minipage}
    \hfill
    \begin{minipage}[h]{0.32\linewidth}
        \centering
        \includegraphics[width=\linewidth]
        {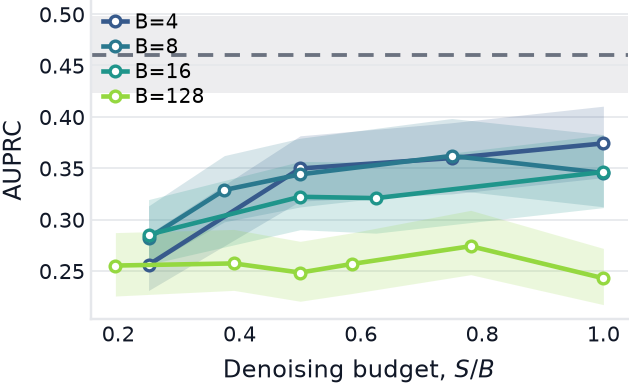}
    \end{minipage}

    \caption{
    \textbf{Left:} Difference in AUPRC between block diffusion and AR under generative rollout and frozen linear probing across four clinical tasks.
    \textbf{Middle:} ICU mortality AUPRC as block size varies under the fixed normalised denoising budget $S/B=0.5$.
    Dashed lines denote the AR baseline. Shaded regions and error bars indicate 95\% bootstrap confidence intervals.
    \textbf{Right:} ICU mortality AUPRC as the normalized denoising budget $S/B$ increases for fixed block sizes.}
    \label{fig:auprc_appendix}
\end{figure}

\subsection{Decoding Interventions}
\label{decoding_intervetions}
\begin{table}[h]
\centering
\caption{ICU mortality decoding strategies for block size $B{=}8$ and $S{=}4$ diffusion steps. Metrics are medians over completed full-validation runs where available; Confidence uses the available full split run. Brackets are $95\%$ CIs over the three random seeds.}
\label{tab:decoding_strategies}
\begin{tabular}{lcc}
\toprule
Strategy & AUROC & AUPRC \\
\midrule
Random & 0.8476 [0.8331, 0.8614] & 0.3440 [0.3123, 0.3792] \\
ReMDM $\eta=0.05$ & 0.8421 [0.8268, 0.8562] & 0.3338 [0.3028, 0.3679] \\
ReMDM $\eta=0.10$ & 0.8446 [0.8307, 0.8590] & 0.3515 [0.3176, 0.3871] \\
ReMDM $\eta=0.20$ & 0.8433 [0.8287, 0.8580] & 0.3474 [0.3158, 0.3800] \\
Confidence & 0.8089 [0.7921, 0.8257] & 0.3487 [0.3154, 0.3844] \\
\bottomrule
\end{tabular}
\end{table}